\documentclass[lettersize, journal]{IEEEtran}
\usepackage{amsmath,amsfonts}
\usepackage{algorithmic}
\usepackage{algorithm}
\usepackage{array}
\usepackage[caption=false,font=normalsize,labelfont=sf,textfont=sf]{subfig}
\usepackage{textcomp}
\usepackage{stfloats}
\usepackage{url}
\usepackage{verbatim}
\usepackage{graphicx}
\usepackage{cite}
\usepackage[utf8]{inputenc} 
\usepackage[T1]{fontenc}    
\usepackage{url}            
\usepackage{booktabs}       
\usepackage{nicefrac}       
\usepackage{microtype}      
\usepackage{xcolor}         
\usepackage{soul}
\usepackage{lipsum}
\definecolor{customyellow}{RGB}{241, 248, 249}
\definecolor{customred}{RGB}{183, 219, 227}

\usepackage{amssymb}
\usepackage{mathtools}
\usepackage{amsthm}
\usepackage{multirow}
\usepackage{colortbl}
\usepackage{adjustbox}
\usepackage{pifont}

\begin{document}

\title{Bidirectional Prototype-Reward co-Evolution for Test-Time Adaptation of Vision-Language Models}

\author{Xiaozhen~Qiao,~Peng~Huang,~Jiakang~Yuan,~Xianda~Guo,~Bowen~Ye,~Chaocan Xue,\\~Ye Zheng,~Zhe Sun,~Xuelong~Li~\IEEEmembership{Fellow,~IEEE} \thanks{

Xiaozhen Qiao is with the School of Information Science and Technology, University of Science and Technology of China, Hefei, 230026, China.(\textit{E-mail:xiaozhennnqiao@mail.ustc.edu.cn})

Peng Huang, Bowen Ye, Chaocan Xue, Ye Zheng, Zhe Sun and Xuelong Li are with the Institute of Artificial Intelligence (TeleAI), China Telecom, P. R. China. (\textit{Corresponding author:~sunzhe@nwpu.edu.cn~xuelong\_li@ieee.org})

Jiakang Yuan is with the College of Future Information Technology, Fudan University, Shanghai, 200433, China

Xianda Guo is with the College of Computer Science, Wuhan University, Wuhan, 430072, China}

}

\markboth{SUBMITTED TO IEEE TRANSACTIONS ON MULTIMEDIA}%
{Shell \MakeLowercase{\textit{et al.}}: A Sample Article Using IEEEtran.cls for IEEE Journals}

\maketitle

\begin{abstract}
Test-time adaptation (TTA) is crucial in maintaining performance of Vision Language Models (VLMs) when facing distribution shifts, particularly when the source data or target labels are inaccessible. Existing TTA methods predominantly leverage the output probability distribution of CLIP for feature evaluation, resulting in biases under domain shifts, which cause misclassified features due to text priors or incorrect textual associations. To address these issues, we propose \underline{B}idirectional Prototype-Reward co-Evolution (BPRE), a novel VLMs framework with TTA that integrates feature quality assessment with prototype evolution via a synergistic feedback loop. First, the Multi-dimensional Quality-aware Reward Module (MQRM) is designed to evaluate feature quality and guide prototype refinement precisely. The continuous refinement of prototype quality via Prototype-Reward Interactive Evolution (PRIE) enhances the computation more robust. Through this bidirectional interaction, the precision of rewards and prototype evolution mutually reinforce each other, forming a self-evolving feedback cycle. Extensive experiments conducted on 15 diverse recognition datasets demonstrate that our model consistently achieves superior performance compared to other SOTA methods, and advances VLM generalization capabilities through emphasizing comprehensive feature evaluation.

\end{abstract}

\begin{IEEEkeywords}
Prototype Learning, Test-Time Adaptation, Visual Language Models.
\end{IEEEkeywords}

\section{Introduction}

\IEEEPARstart{R}{ecent} advances in multimodal models~\cite{chen2023adaptive, zhang2024integrating, yin2025structure, yin_grpose, yao2025bi} have opened new opportunities to integrate human language into various computer vision tasks. VLMs such as CLIP~\cite{radford2021learning} demonstrate remarkable zero-shot generalization capabilities, enabling them to perform well on tasks without task-specific training. Their performance drops significantly when faced with domain shifts. This decline is mainly due to discrepancies between the data distributions of the source and target domains. Moreover, such domain discrepancies are inevitable in dynamic and ever-changing environments, making effective domain adaptation for VLMs an urgent and challenging problem. 

\begin{figure}[!t]
 \centering
\includegraphics[width=1.00\columnwidth]{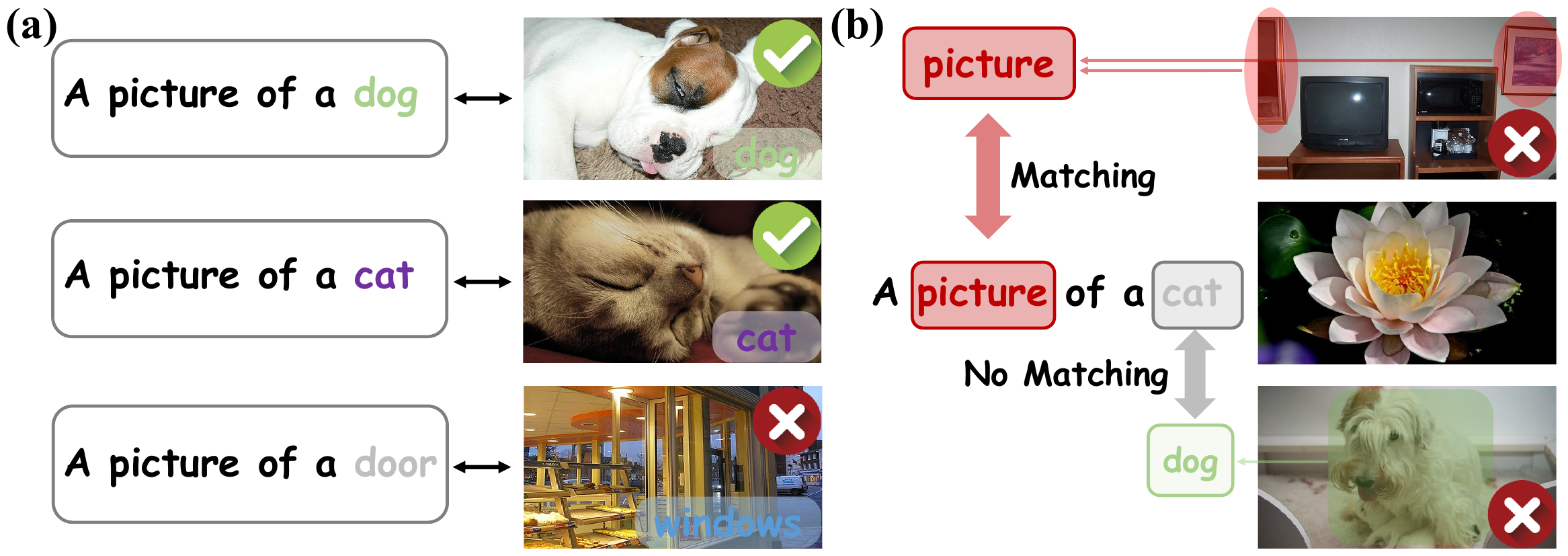}
\caption{(a) Images of dogs or cats may lack visual details, but they can still be prioritized due to their strong textual representations. In contrast, visually rich samples, like a window, may be misclassified as the nearest known category, such as a door, if their textual alignment is weak.This happens when their textual alignment is weak. (b) Certain image regions may be visually associated with the term “picture.” These regions can be erroneously emphasized during feature evaluation.}
\label{fig:2}  
\vspace{-15pt}
\end{figure}

Efficient adaptation methods~\cite{zhang2022tip, zhu2023prompt, chen2024comkd, wang2025decoupled} have been proposed to fine-tune VLMs using only a limited number of target domain samples. The requirement for target domain annotations remains a critical bottleneck, especially in dynamic environments with frequent distribution shifts~\cite{recht2019imagenet, zhang2024hiker}. In addition, several studies have explored the test-time adaptation (TTA) for VLMs, aiming to efficiently optimize the model during the test phase with minimal cost. Shu \textit{et al.}~\cite{shu2022test} proposed test-time prompt tuning (TPT), which optimizes prompts for individual test samples to improve out-of-distribution generalization. This framework was further enhanced by DiffTPT~\cite{feng2023diverse}, which incorporates diffusion-based augmentations. In addition, Karmanov \textit{et al.}~\cite{karmanov2024efficient} introduced a training-free approach based on dynamic visual caches. Zhang \textit{et al.}~\cite{zhang2024dual} developed a dual prototype evolving strategy that accumulates task-specific knowledge from both visual and textual modalities during testing.

Most existing TTA methods~\cite{zhang2022tip, karmanov2024efficient, zhang2024dual} emphasize leveraging historical feature information to accumulate class-specific task knowledge via caches and prototypes. These approaches construct accurate and discriminative target class representations, which are critical for enhancing model robustness to distribution shifts and improving efficiency during test time. Current methods~\cite{karmanov2024efficient, zhang2024dual} primarily evaluate feature quality using CLIP's output probability distribution, which fuses text and image features, and inherently uses the text modality as a reference. However, distribution shifts during testing resulting in misalignment between text and image embedding~\cite{vosoughi2024cross}, leading to biases in feature selection and evaluation, as illustrated in Fig.~\ref{fig:2}. On the one hand, text priors drive the model to prioritize features aligning strongly with pre-trained text descriptions, even when such features lack adequate visual information, while misclassifying visually rich samples with weak text representations (Fig.~\ref{fig:2}(a)). On the other hand, misalignment of text descriptions introduces feature evaluation biases, where CLIP incorrectly associates generalized text descriptions (\textit{e.g.},``picture'') with certain image features, leading the model to mistakenly consider them high-quality samples while overlooking their true semantics, thereby affecting accuracy of feature selection (Fig.~\ref{fig:2}(b)).

\begin{figure*}[!t]
    \centering
    \includegraphics[width=\textwidth]{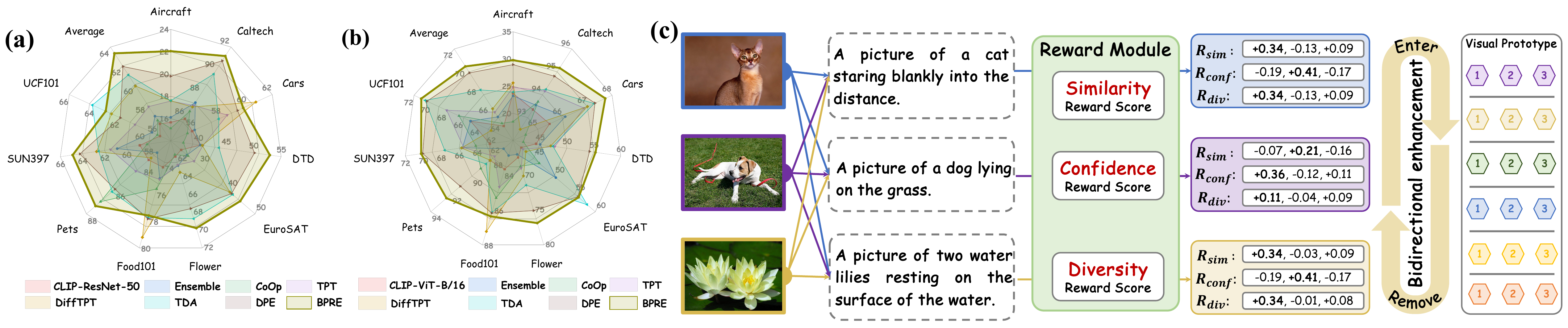}
    \caption{A comparative analysis of cross-dataset generalization performance between CLIP models pre-trained with ResNet-50 (a) and ViT-B/16 (b) image encoders. BPRE demonstrates competitive results across ten diverse datasets, showcasing its robust generalization capabilities. (c) Bidirectional Prototype-Reward co-Evolution diagram.}
    \label{fig:1}
    \vspace{-15pt}
\end{figure*}

Inspired by these insights, we propose a novel framework for test-time adaptation of VLMs, named \textbf{B}idirectional \textbf{P}rototype-\textbf{R}eward co-\textbf{E}volution (BPRE). BPRE iteratively refines task-specific knowledge, as illustrated in Fig.~\ref{fig:1}(c). In contrast to existing methods, we introduce the \textbf{M}ulti-dimensional \textbf{Q}uality-aware \textbf{R}eward \textbf{M}odule (MQRM), which evaluates features by incorporating intrinsic image properties beyond mere textual correspondence. Integrating visual information with semantic similarity and feature diversity, this module enables comprehensive image feature quality assessment, thereby mitigating excessive reliance on text modality. Semantic similarity ensures that selected features align with the task-specific semantic space, reducing errors caused by text biases. Meanwhile, feature diversity broadens coverage across varied data distributions, preventing overfitting and improving prototype selection. In addition, the \textbf{P}rototype-\textbf{R}eward \textbf{I}nteractive \textbf{E}volution(PRIE) mechanism adaptively updates historical features and Prototypes based on reward scores. This bidirectional reinforcement loop allows refined prototypes to improve reward estimation, creating a self-reinforcing optimization process that boosts noise robustness, promotes interpretable cross-modal learning, and yields more discriminative and stable representations. The test-time generalization of BPRE is evaluated on 15 diverse recognition datasets under natural distribution shifts and cross-dataset scenarios. As shown in Fig.~\ref{fig:1}(a)(b), BPRE achieves competitive cross-dataset generalization performance with both ResNet-50 and ViT-B/16 architectures.

Our main contributions are summarized as follows:

\begin{itemize}
        \item We introduce the Multi-dimensional Quality-aware Reward Module (MQRM), integrating similarity, confidence, and diversity reward scores for comprehensive feature evaluation to improve prototype selection.
        \item We propose the Prototype-Reward Interactive Evolution(PRIE) Module, establishing a bidirectional reinforcement process where reward scores guide prototype evolution, and refined prototypes enhance reward estimation.
        \item Experiments on 15 diverse image recognition datasets demonstrate that our method achieves competitive generalization performance and advances the generalization capabilities of VLMs.
        \end{itemize}

\begin{figure*}[!t]
 \centering
\includegraphics[width=1.0\linewidth]{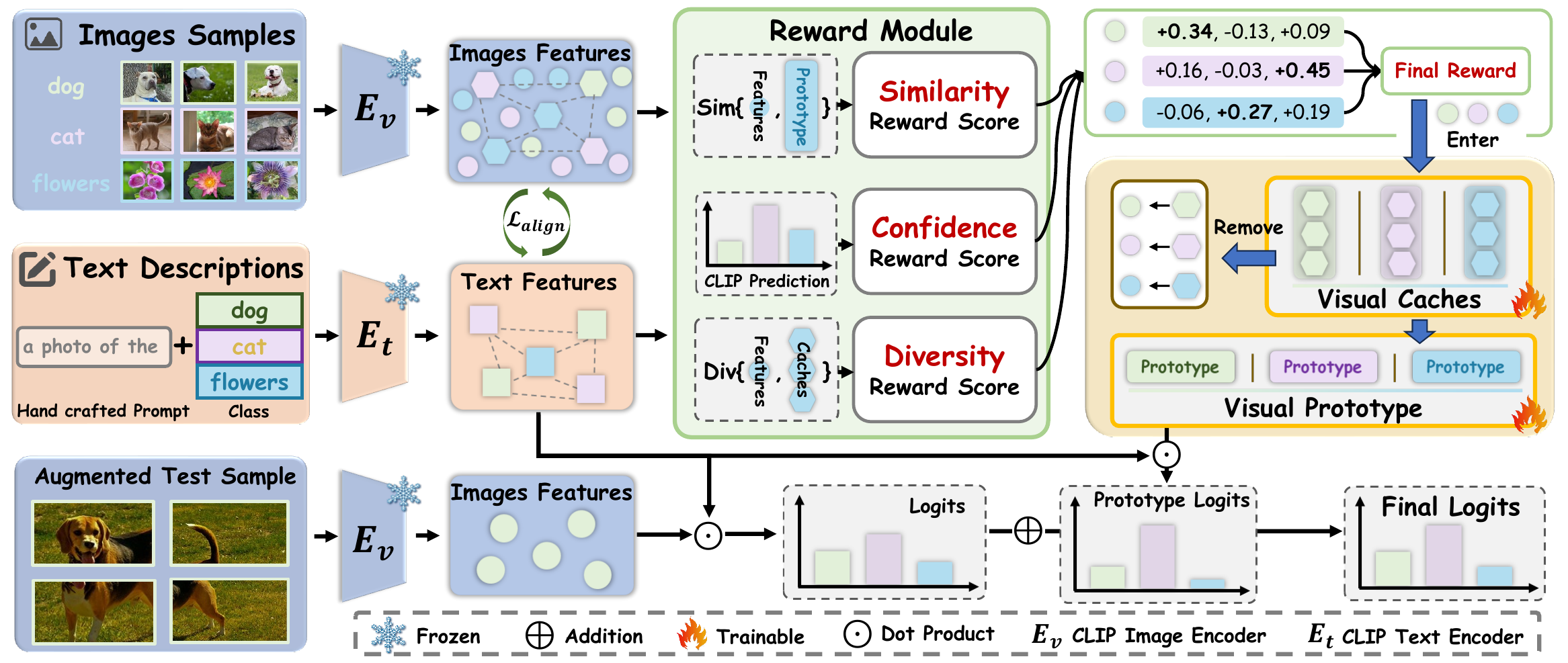}
\caption{Overview of our proposed BPRE framework. Given text and image inputs, encoders $E_t(\cdot)$ and $E_v(\cdot)$ extract text features $f_t$ and visual features $f_v$. A probability distribution $\mathbb{P}_{CLIP}(y=y_c|X_{\mathtt{test}})$ is computed based on these features. A reward module then integrates similarity, confidence, and diversity metrics to derive a composite reward score $R_{final}$, guiding prototype updates. Finally, a residual-based refinement step adjusts the logits by compensating for the residuals between updated prototypes and test samples.}
\label{fig:pipeline}  
\vspace{-10pt}
\end{figure*}

\section{Related Works}
\subsection{Vision-Language Models}
Large-scale VLMs such as CLIP~\cite{radford2021learning} and SigLIP~\cite{zhai2023sigmoid} have demonstrated remarkable success in visual understanding tasks by leveraging massive image-text pre-training~\cite{du2022survey, zhang2024vision,lu2025omnicaptioner}. Their strong zero-shot transferability makes them effective backbones for tasks like image recognition~\cite{hegde2023clip, liu2024remoteclip} and object detection~\cite{wu2023cora, lei2024ez}.
To better adapt VLMs to downstream tasks, researchers have proposed two main categories of methods: prompt learning methods~\cite{zhou2022learning, zhou2022conditional, khattak2023maple, zhu2023prompt, chen2024comkd} and adapter-based methods~\cite{zhang2022tip, yu2023task, gao2024clip, zhou2024dynamic}. Prompt learning methods like CoOp~\cite{zhou2022learning} and CoCoOp~\cite{zhou2022conditional} focus on learning continuous prompt vectors with few-shot labeled data to bridge the domain gap between pre-training and downstream tasks. These methods achieve impressive performance but require labeled training samples that are often unavailable in real-world scenarios. Despite the success of adapter-based methods like Tip-Adapter~\cite{zhang2022tip} and TaskRes~\cite{yu2023task} in enhancing visual-textual representations, their dependence on labeled target data remains a critical bottleneck. Here we present a test-time adaptation framework that overcomes this limitation by enabling real-time model adaptation without prior exposure to target domain labels or training samples. Our approach facilitates robust model deployment across diverse, unfamiliar environments where labeled data collection is unfeasible.

\subsection{Test-Time Adaptation}
Test-time adaptation addresses domain shift challenges by enabling dynamic model adaptation during deployment using unlabeled test samples~\cite{tang2023neuro, karmanov2024efficient, zhang2024dual,yuan2024reg}. This approach has proven effective across various computer vision tasks, enhancing model robustness and maintaining performance when encountering out-of-distribution scenarios in real-world applications.
Test-time adaptation of VLMs has recently gained significant research attention. TPT~\cite{shu2022test} pioneered this direction by leveraging consistency across augmented views of test samples. The following works enhanced this approach: DiffTPT~\cite{feng2023diverse} introduced diffusion-based augmentations, while C-TPT~\cite{yoon2024c} focused on addressing calibration errors. Moving beyond single-sample adaptation, TDA~\cite{karmanov2024efficient} and DMN~\cite{zhang2024dual0} introduced memory-based mechanisms to leverage information across multiple test samples, marking a shift from instance-level to collective adaptation strategies. DPE~\cite{zhang2024dual} proposes a dual-modality prototype evolution mechanism, which achieves more accurate feature modeling through dynamic updating and optimization of dual-modality feature representations during testing. However, existing methods, including DPE~\cite{zhang2024dual}, focus solely on prototype evolution, overlooking the mutual interaction between prototypes and sample quality. We address this by proposing a \textbf{B}idirectional \textbf{P}rototype-\textbf{R}eward co-\textbf{E}nvolution framework that enables quality-aware prototype evolution while ensuring consistent feature representation through mutual guidance.

\section{Method}

As shown in Fig.~\ref{fig:pipeline}, we propose the BPRE framework to enhance CLIP’s zero-shot generalization at test time. Unlike prior methods~\cite{zhang2022tip,karmanov2024efficient,zhang2024dual}, BPRE introduces a \textbf{Bidirectional Prototype-Reward co-Evolution mechanism}, where prototypes and rewards mutually refine each other. This enables quality-aware adaptation and consistent feature representation, dynamically updating prototypes with unlabeled test data \(\mathcal{D}_{\mathtt{test}}\) without supervision.

\subsection{Preliminary}

\subsubsection{Zero-Shot CLIP}
CLIP~\cite{radford2021learning} consists of dual encoders: a visual encoder $\mathcal{E}_v(\cdot)$ and a text encoder $\mathcal{E}_t(\cdot)$, which project images and text into a shared $d$-dimensional embedding space. CLIP performs zero-shot prediction for a $C$-class classification task by measuring similarities between image and text embeddings. Given a test image $X_{\mathtt{test}} \in \mathcal{D}_{\mathtt{test}}$ and class-specific text descriptions ${\mathcal{T}_c}$, the prediction process can be formulated as:

\begin{equation}
\mathbb{P}_{\mathrm{CLIP}}(y = y_c \mid X_{\mathtt{test}}) = 
\frac{\exp\left( \mathrm{sim}(\mathcal{E}_t(\mathcal{T}_c), \mathcal{E}_v(X_{\mathtt{test}})) / \tau \right)}
     {\sum_{t^{\prime}} \exp\left( \mathrm{sim}(f_{t^{\prime}}, \mathcal{E}_v(X_{\mathtt{test}})) / \tau \right)},
\label{eq:clip_prob}
\end{equation}

\noindent where $\mathrm{sim}(\cdot,\cdot)$ denotes cosine similarity and $\tau$ is a temperature scaling parameter.

\subsubsection{Test-Time Prompt Tuning} TPT~\cite{shu2022test} improves CLIP's zero-shot generalization by learning adaptive prompts from test samples. For each test image $X_{\mathtt{test}}$, TPT generates $N$ augmented views ${\mathcal{A}_n(X_{\mathtt{test}})}_{n=1}^N$ and aggregates their predictions based on confidence. Specifically, TPT selects predictions with entropy below threshold $t$ and computes their average within the top $\rho$-percentile:

\begin{align}
\mathbb{P}_{\mathtt{TPT}}(X_{\mathtt{test}}) &= \frac{1}{\rho N} \sum_{n=1}^{N} \mathbf{1}\Big[\mathcal{H}\big(\mathbb{P}_{\mathtt{CLIP}}(\mathcal{A}_n(X_{\mathtt{test}}))\big) \leq t\Big] \nonumber \\
&\quad \times \mathbb{P}_{\mathtt{CLIP}}(\mathcal{A}_n(X_{\mathtt{test}}))
\end{align}

where $\mathcal{H}(p) = -\sum_{i=1}^C p_i \log p_i$ is the prediction entropy. The prompt is optimized to minimize the entropy of the final prediction: $\min \mathcal{H}(\mathbb{P}_{\mathtt{TPT}}(X_{\mathtt{test}}))$.

\subsection{Multi-dimensional Quality-aware Reward Module}
To effectively assess samples' contributions to model performance and guide feature learning, we introduce a Multi-dimensional Quality-aware Reward Module(MQRM), which dynamically assigns quality scores by integrating multiple characteristics.

\subsubsection{Similarity Reward Score}
In cross-modal learning, semantic consistency between sample features and category prototypes serves as a crucial indicator for sample quality evaluation. To this end, we design a similarity-based reward $R_{sim}$ that measures the alignment between sample features and prototype representations:
\begin{equation}
R_{sim} = \frac{1}{|\mathcal{P}|} \sum_{p_k \in \mathcal{P}} \cos(\frac{f_i}{\|f_i\|_2}, \frac{p_k}{\|p_k\|_2}),
\label{equation:sim}
\end{equation}
where $f_i \in \mathbb{R}^d$ is the feature vector of sample $x_i$, and $\mathcal{P} = \{p_1, p_2, \dots, p_K\}$ denotes the set of category prototypes, with each $p_k \in \mathbb{R}^d$ representing the prototype of category $k$. The function $\cos(\cdot, \cdot)$ computes the cosine similarity between the normalized feature vector $f_i$ and prototype $p_k$, while $\|\cdot\|_2$ denotes the $L_2$-norm for normalization. This similarity reward enhances feature discriminability by enforcing semantic consistency and aligning prototype evolution with category semantics.

\begin{figure*}[t]
  \centering
  \includegraphics[width=0.95\textwidth]{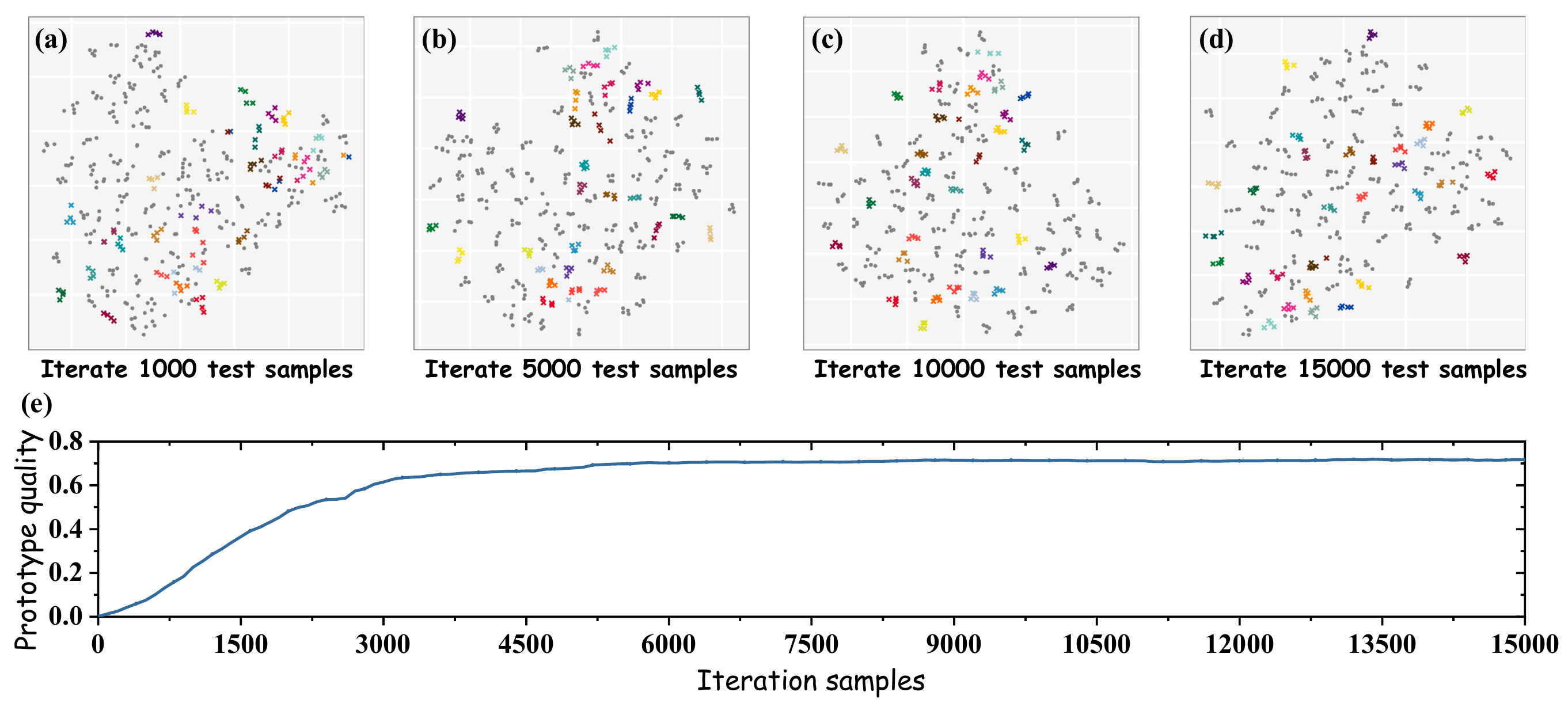}
  \caption{(a-d) t-SNE~\cite{van2008visualizing} visualizations of image features in the priority queues demonstrate that as additional samples are incorporated, class-specific features become more tightly clustered on the Food101~\cite{bossard2014food} dataset. (e) As the number of input samples grows, the reward score computed by the Multi-Dimensional Quality-Aware Reward Module increases.}
  \label{fig:3}
  \vspace{-15pt}  
\end{figure*}


\subsubsection{Confidence Reward Score}
To assess the model's prediction confidence, we define a confidence reward $R_{conf}$ based on the entropy of the predicted probability distribution. Utilizing temperature-scaled normalization, $R_{conf}$ is derived from CLIP's predicted probability distribution $\mathbf{P}_{CLIP}$:
\begin{equation}
R_{conf} =  1 - \frac{-\sum_{i=1}^{C} \mathbb{P}_{CLIP}(X_{test}) \log \mathbb{P}_{CLIP}(X_{test})}{\log C},
\label{equation:conf}
\end{equation}
where $C$ is the number of categories. This entropy-based confidence reward prioritizes high-certainty predictions, enhancing model reliability while fostering more robust feature representations.

\subsubsection{Diversity Reward Score}
To prevent representation collapse and encourage diverse feature exploration, we introduce a diversity reward $R_{div}$ based on historical feature comparisons. We maintain a fixed-size feature memory queue $\mathcal{H} = \{h_1, h_2, \dots, h_M\}$, where each $h_j \in \mathbb{R}^d$ represents the average feature of a past batch, and $M$ is the queue size. For a given sample feature $f_i$, its novelty is measured by the maximum cosine similarity with historical features:
\begin{equation}
S_{max} = \max_{h_j \in \mathcal{H}} \cos(\frac{f_i}{\|f_i\|_2}, \frac{h_j}{\|h_j\|_2}),
\label{equation:max}
\end{equation}
The diversity reward is then defined as:
\begin{equation}
R_{div} = 1 - S_{max}.
\label{equation:div}
\end{equation}
This mechanism enhances feature diversity while maintaining discriminability, improving the model’s generalization across downstream tasks.

By integrating the proposed reward components, we define an adaptive comprehensive reward function that evaluates sample quality while guiding prototype evolution:
\begin{equation}
R = \lambda_{sim} R_{sim} + \lambda_{conf} R_{conf} + \lambda_{div} R_{div},
\label{equation:reward_1}
\end{equation}
where $\lambda_{sim}$, $\lambda_{conf}$, and $\lambda_{div}$ are hyperparameters that balance the contributions of similarity reward $R_{sim}$, confidence reward $R_{conf}$, and diversity reward $R_{div}$, respectively. 
\begin{equation}
R_{final} = R_{min} + (R - R_{min}) * \eta_{\text{warm}},
\label{equation:reward2}
\end{equation}
where $R_{min}$ is the minimum reward threshold, and $\eta_{\text{warm}}$ is a warmup factor. Initially, when $\eta_{\text{warm}}$ is small, $R_{final}$ remains close to $R_{min}$, stabilizing early adaptation. As $\eta_{\text{warm}}$ increases, $R_{final}$ gradually approaches $R$, enhancing adaptation while mitigating fluctuations.


\subsection{Prototype-Reward Interactive Evolution}
Building on our MQRM, we propose a dynamic prototype evolution mechanism that weights feature contributions based on their computed rewards. This approach adaptively refines category-specific prototypes, creating a more robust and discriminative feature space for cross-modal recognition.

\subsubsection{Prototype Representation and Initialization}
For a $C$-category classification task, we maintain prototype representations $v_c \in \mathbb{R}^d$ in the image feature space. Initially, the visual prototype $v_c^{(0)}$ for category $c$ is set as:

\begin{equation}
v_c^{(0)} = \frac{1}{|\mathcal{D}_c|} \sum_{x_{i} \in \mathcal{D}_c} f_v,
\label{equation:prototype_0}
\end{equation}
where $\mathcal{D}_c$ denotes the set of samples belonging to category $c$.

\begin{table*}[t]
\renewcommand\arraystretch{1}
\renewcommand{\tabcolsep}{2pt}
\vspace{1pt}
\centering
\caption{Robustness Evaluation under Natural Distribution Shifts. We report top-1 accuracy (\%) across methods using CLIP's ResNet-50 and ViT-B/16 backbones. \textbf{Bold} and \underline{underlined} values indicate the best and second-best performance, respectively.}
\resizebox{1.00\linewidth}{!}{
\begin{tabular}{lrccccccc}
\toprule
{Method} & Source& ImageNet & ImageNet-A & ImageNet-V2 & ImageNet-R & ImageNet-S & {Average} & {OOD Average} \\
\midrule
CLIP-ResNet-50~\cite{radford2021learning} & ICML'21 & 58.16 & 21.83 & 51.41 & 56.15 & 33.37 & 44.18 & 40.69 \\ 
\midrule
CoOp~\cite{zhou2022learning} & IJCV'22 & 63.33 & 23.06 & 55.40 & 56.60 & 34.67 & 46.61 & 42.43 \\
CoCoOp~\cite{zhou2022conditional} & CVPR'22 & 62.81 & 23.32 & 55.72 & 57.74 & 34.48 & 46.81 & 42.82 \\
Tip-Adapter~\cite{zhang2022tip} & ECCV'22 & 62.03 & 23.13 & 53.97 & 60.35 & 35.74 & 47.04 & 43.30 \\

\midrule
TPT~\cite{shu2022test} & NeurIPS'22  & 60.74 & 26.67 & 54.70 & 59.11 & 35.09 & 47.26 & 43.89 \\
DiffTPT~\cite{feng2023diverse} &  ICCV'23 & 60.80 & \textbf{31.06} & 55.80 & 58.80 & 37.10 & 48.71 & 45.69 \\
TDA~\cite{karmanov2024efficient} &  CVPR'24  & 61.35 & 30.29 & 55.54 & 62.58 & 38.12 & 49.58 & 46.63 \\ 

DMN-ZS~\cite{zhang2024dual0} & CVPR'24  &  \textbf{63.87} & 28.57 & 56.12 & 61.44 & 39.84 & 49.97 & 46.49 \\ 
DPE~\cite{zhang2024dual} & NeurIPS'24  &  63.41 & 30.15 & \underline{56.72} & \underline{63.72} & \textbf{40.03} & \underline{50.81} & \underline{47.66} \\  
TPS~\cite{sui2024just} & WACV'25  & 61.47 & 30.48 & 54.96 & 62.87 & 37.14 & 49.38 & 46.36 \\
\textbf{BPRE} & \textbf{Ours} & \underline{63.77} & \underline{30.77} & \textbf{56.97} & \textbf{63.94} & \underline{39.91} & \textbf{51.07} & \textbf{47.90} \\
\midrule
\midrule
CLIP-ViT-B/16~\cite{radford2021learning} & ICML'21  & 66.73 & 47.87 & 60.86 & 73.98 & 46.09 & 59.11 & 57.20 \\
\midrule
CoOp~\cite{zhou2022learning} & IJCV'22 & 71.51 & 49.71 & 64.20 & 75.21 & 47.99 & 61.72 & 59.28 \\
CoCoOp~\cite{zhou2022conditional} & CVPR'22 & 71.02 & 50.63 & 64.07 & 76.18 & 48.75 & 62.13 & 59.91 \\
Tip-Adapter~\cite{zhang2022tip} & ECCV'22 & 70.75 & 51.04 & 63.41 & 77.76 & 48.88 & 62.37 & 60.27 \\
\midrule
TPT~\cite{shu2022test} &NeurIPS'22& 68.98 & 54.77 & 63.45 & 77.06 & 47.94 & 62.44 & 60.81 \\
DiffTPT~\cite{feng2023diverse} & ICCV'23 & 70.30 & 55.68 & 65.10 & 75.00 & 46.80 & 62.28 & 60.52 \\
TDA~\cite{karmanov2024efficient} & CVPR'24 & 69.51 & \underline{60.11} & 64.67 & 80.24 & 50.54 & 65.01 & 63.89 \\ 
DMN-ZS~\cite{zhang2024dual0} &CVPR'24& \textbf{72.25} & 58.28 & 65.17 & 78.55 & \underline{53.20} & 65.49 & 63.80 \\
DPE~\cite{zhang2024dual} &NeurIPS'24& 71.91 & 59.63 & \underline{65.44} & \underline{80.40} & 52.26 & \underline{65.93} & \underline{64.43} \\  
TPS~\cite{sui2024just}& WACV'25 & 70.19 & 60.08 & 64.73 & 80.27 & 49.95 & 65.04 & 63.76 \\
\textbf{BPRE} & \textbf{Ours} & \underline{72.22} & \textbf{60.67} & \textbf{65.50} & \textbf{80.68} & \textbf{53.45} & \textbf{66.50} & \textbf{65.08} \\
\bottomrule
\end{tabular}
}

\label{tab:ood-main}
\end{table*}

\subsubsection{Quality-Aware Visual Prototype Evolution}
Leveraging the comprehensive quality assessment derived from our multi-dimensional reward mechanism, we propose a quality-aware momentum update strategy for dynamic prototype evolution. This mechanism strategically incorporates sample contributions by utilizing their reward scores $R$ as quality indicators, thereby ensuring that samples with higher reward values exert greater influence on the prototype evolution process. Specifically, for category $c$, the visual prototype update at time step $t$ is formulated as:
\begin{equation}
v_c^{(t+1)} = m \cdot v_c^{(t)} + (1 - m) \cdot \sum_{x_i \in \mathcal{D}_c} w_i^{(t)} \cdot f_v,
\label{equation:prototype}
\end{equation}
where $m \in [0,1]$ is a momentum coefficient that stabilizes the update process by controlling the balance between historical information and new observations. The adaptive weight $w_i^{(t)}$ for each sample is computed through temperature-scaled softmax normalization:
\begin{equation}
w_i^{(t)} = \frac{\exp(R_{final} / \tau)}{\sum_{j \in \mathcal{D}_c} \exp(R_{final} / \tau)}.
\label{equation:reward_3}
\end{equation}
Here, $R_{final}$ is the comprehensive quality score from our multi-dimensional reward mechanism, and $\tau$ controls the weight distribution's sharpness.

Overall, our dual promotion mechanism establishes a dynamic feedback loop between reward estimation and prototype evolution. This operates through two complementary processes: high-fidelity prototypes enable precise reward assessment, while refined reward signals guide targeted prototype updates. As shown in Fig.~\ref{fig:3}, increasing iterative samples leads to tighter class-specific feature clustering and higher reward scores. This co-evolution progressively enhances the framework's ability to model visual-semantic relationships and adapt to intra-class variations, improving robustness against noise and fostering interpretable cross-modal feature learning for more discriminative representations.

\begin{table*}[t]
\renewcommand\arraystretch{1}
\renewcommand{\tabcolsep}{2pt}
  \vspace{5pt}
  \centering
  \caption{Robustness Evaluation under Natural Distribution Shifts. We report top-1 accuracy (\%) across methods using CLIP's ResNet-50 and ViT-B/16 backbones. \textbf{Bold} and \underline{underlined} values indicate the best and second-best performance, respectively.}
  \resizebox{1.00\linewidth}{!}{
    \begin{tabular}{lr p{1.2cm}<{\centering}p{1.2cm}<{\centering}p{1.2cm}<{\centering}p{1.2cm}<{\centering}p{1.2cm}<{\centering}p{1.2cm}<{\centering}p{1.2cm}<{\centering}p{1.2cm}<{\centering}p{1.2cm}<{\centering}p{1.2cm}<{\centering}p{1.2cm}<{\centering}}
      \toprule
      Method & Source & Aircraft & Caltech & Cars & DTD & EuroSAT & Flower & Food101 & Pets & SUN397 & UCF101 & Average \\
      \midrule
      CLIP-ResNet50~\cite{radford2021learning}& ICML'21&15.66 & 85.88 & 55.70 & 40.37 & 23.69 & 61.75 & 73.97 & 83.57 & 58.80 & 58.84 & 55.82 \\
      \midrule
      CoOp~\cite{zhou2022learning} & IJCV'22 & 15.12 & 86.53 & 55.32 & 37.29 & 26.20 & 61.55 & 75.59 & 87.00 & 58.15 & 59.05 & 56.18 \\
      CoCoOp~\cite{zhou2022conditional} & CVPR'22 & 14.61 & 87.38 & 56.22 & 38.53 & 28.73 & 65.57 & 76.20 & 88.39 & 59.61 & 57.10 & 57.23 \\
      Tip-Adapter~\cite{zhang2022tip} & ECCV'22 & 16.11 & 87.26 & 55.89 & 40.37 & 25.79 & 62.77 & 74.82 & 82.97 & 60.85 & 59.48 & 56.63 \\
      
      \midrule
      TPT~\cite{shu2022test} & NeurIPS'22 & 17.58 & 87.02 & 58.46 & 40.84 & 28.33 & 62.69 & 74.88 & 84.49 & 61.46 & 60.82 & 57.66 \\
      DiffTPT~\cite{feng2023diverse}& ICCV'22 & 17.60 & 86.89 & \textbf{60.71} & 40.72 & 41.04 & 63.53 & \textbf{79.21} & 83.40 & 62.72 & 62.67 & 59.85 \\
      TDA~\cite{karmanov2024efficient} & CVPR'24 & 17.61 & 89.70 & 57.78 & 43.74 & \underline{42.11} & \underline{68.74} & 77.75 & \underline{86.18} & 62.53 & \textbf{64.18}  & 61.03\\ 
      DPE~\cite{zhang2024dual}& NeurIPS'24 & \underline{19.80} & \underline{90.83} & 59.26 & \underline{50.18} & 41.67 & 67.60 &\underline{77.83} & 85.97 & \underline{64.23} & 61.98 & \underline{61.93} \\  
      \textbf{BPRE}& Ours&\textbf{22.05} &\textbf{91.24} & \underline{59.71} & \textbf{52.96} & \textbf{44.74} & \textbf{69.79} &  77.59 & \textbf{87.35} & \textbf{64.88} & \underline{63.13} & \textbf{63.34} \\
      \midrule
      \midrule
      CLIP-ViT-B/16 &ICML'21 & 23.67 & 93.35 & 65.48 & 44.27 & 42.01 & 67.44 & 83.65 & 88.25 & 62.59 & 65.13 & 63.58 \\
      \midrule
      CoOp~\cite{zhou2022learning} & IJCV'22 & 18.47 & 93.70 & 64.51 & 41.92 & 46.39 & 68.71 & 85.30 & 89.14 & 64.15 & 66.55 & 63.88 \\
      CoCoOp~\cite{zhou2022conditional} & CVPR'22  & 22.29 & 93.79 & 64.90 & 45.45 & 39.23 & 70.85 & 83.97 & 90.46 & 66.89 & 68.44 & 64.63 \\
      Tip-Adapter~\cite{zhang2022tip} & ECCV'22 & 16.11 & 87.26 & 55.89 & 40.37 & 25.79 & 62.77 & 74.82 & 82.97 & 60.85 & 59.48 & 56.63 \\      
      \midrule
      TPT~\cite{shu2022test}  & NeurIPS'22 & 24.78 & 94.16 & 66.87 & 47.75 & 42.44 & 68.98 & 84.67 & 87.79 & 65.50 & 68.04 & 65.10 \\
      DiffTPT~\cite{feng2023diverse} & ICCV'23 & 25.60 & 92.49 & 67.01 & 47.00 & 43.13 & 70.10 & \textbf{87.23} & 88.22 & 65.74 & 62.67 &65.47 \\
      TDA~\cite{karmanov2024efficient} & CVPR'24 & 23.91 & 94.24 & 67.28 & 47.40 &  \textbf{58.00} & 71.42 & 86.14 & 88.63 & 67.62 &\underline{70.66} & 67.53 \\ 
      DPE~\cite{zhang2024dual} & NeurIPS'24 &\underline{28.95} & \underline{94.81} & \underline{67.31} &\underline{54.20} & 55.79 & \underline{75.07} & 86.17 &\underline{91.14} &\underline{70.07} & 70.44 &\underline{69.40} \\
      \textbf{BPRE} & Ours & \textbf{29.76} & \textbf{95.17} &  \textbf{67.73} &  \textbf{55.32} &\underline{55.94} & \textbf{76.86} & \underline{86.36} & \textbf{92.48} & \textbf{70.24} &  \textbf{71.13} & \textbf{70.10} \\
      \bottomrule
    \end{tabular}
  } 
  \label{tab:fine-grained}
\end{table*}

\section{Experiments}
We evaluate BPRE using a comprehensive evaluation protocol that emphasizes distributional robustness and cross-dataset generalization. Furthermore, systematic ablation studies are carried out to validate the efficacy of our proposed design principles.

\subsection{Datasets}
Our evaluation protocol encompasses two primary benchmarking scenarios: distribution shift robustness and cross-domain generalization. The distribution robustness assessment utilizes ImageNet~\cite{deng2009imagenet} and its variants (ImageNet-A~\cite{hendrycks2021natural}, -V2~\cite{recht2019imagenet}, -R~\cite{hendrycks2021many}, and -Sketch~\cite{wang2019learning}). For cross-domain evaluation, we employ a diverse suite of 10 recognition datasets, spanning fine-grained categories (FGVCAircraft~\cite{maji2013fine}, StanfordCars~\cite{krause20133d}), texture (DTD~\cite{cimpoi2014describing}), natural objects (Caltech101~\cite{fei2004learning}, Flowers102~\cite{nilsback2008automated}, Food101~\cite{bossard2014food}, OxfordPets~\cite{parkhi2012cats}), scenes (SUN397~\cite{xiao2010sun}), satellite imagery (EuroSAT~\cite{helber2019eurosat}), and action recognition (UCF101~\cite{soomro2012ucf101}).

We implement our approach using PyTorch with CLIP (ResNet-50 and ViT-B/16 architecture). Following TPT, each test image was augmented into 64 views (63 augmented + 1 original). The dynamic prototype evolution utilized a momentum-based update mechanism (momentum = 0.9) with our Multi-dimensional Quality-aware Reward Module. For prototype evolution, we set temperature $\tau=0.01$. Our cache maintains 3 samples per class with a normalized entropy threshold of 0.1. During testing, the reward calculator underwent a 1000-step warmup with a minimum threshold of 0.1, while prototype updates occur every 10 iterations through weighted feature averaging. Experiments were conducted on a single NVIDIA RTX 4090, and we adopted t-SNE (perplexity 30) to project prototype features into a 2D space.

\begin{table}[t]
\normalsize
\begin{center}

\caption{Efficiency comparison on ImageNet~\cite{deng2009imagenet}. We reported testing time, achieved accuracy, and performance gains compared to zero-shot CLIP model.}

\begin{tabular}{lccc}
\toprule
Method  & Time & Acc & Gain \\ 
\midrule
CLIP~\cite{radford2021learning}  & 11 min  &  59.81&  -\\
TPT~\cite{shu2022test} &   $>$ 10 h &  60.74  & +0.93  \\
DiffTPT~\cite{feng2023diverse} &  $>$ 20 h &  60.80  & +0.99  \\
TDA~\cite{karmanov2024efficient} & 32 min  &  61.35  & +1.54  \\
DPE~\cite{zhang2024dual} &  2 h 37 min   & 63.41 & +3.60 \\
\textbf{BPRE}(ours) &  3 h 04 min   &  \textbf{63.77}  & \textbf{+3.96}  \\
\bottomrule
\end{tabular}
\end{center}

\label{table:efficiency}
\vspace{-20pt}
\end{table}

\subsection{Comparison with SOTA Methods} 

To validate our approach, we compared several state-of-the-art TTA methods for CLIP~\cite{radford2021learning}. We included the zero-shot performance of CLIP using both single-prompt and prompt-ensemble strategies. Our analysis encompassed prompt optimization methods (TPT~\cite{shu2022test}, including DiffTPT~\cite{feng2023diverse}), cache-based approaches (TDA~\cite{karmanov2024efficient}, DMN-ZS~\cite{zhang2024dual0}), and prototype-focused methods (TPS~\cite{sui2024just}, DPE~\cite{zhang2024dual}). It is noted that all comparison results are directly cited from their original publications.

\subsubsection{Robustness to Distribution Shifts} 


To evaluate the robustness of our method under natural distribution shifts, we conducted experiments on ImageNet and four challenging out-of-distribution (OOD) variants. As summarized in Tab.~\ref{tab:ood-main}, BPRE consistently outperformed existing approaches across both ResNet-50 and ViT-B/16 backbones, demonstrating strong robustness to domain shifts. Specifically, while baseline CLIP~\cite{radford2021learning} achieved reasonable in-domain accuracy on ImageNet, its performance degraded sharply when evaluated on OOD datasets such as ImageNet-A and ImageNet-Sketch, highlighting its vulnerability to domain shifts. Several methods, such as CoOp~\cite{zhou2022learning} or CoCoOp~\cite{zhou2022conditional}, provided only moderate improvements, indicating that prompt learning alone is insufficient for robust test-time adaptation, especially in the absence of target domain labels. Among recent test-time adaptation strategies, BPRE achieved superior performance on both architectures. For the ResNet-50 backbone, BPRE surpasses prompt-tuning approaches (TPT~\cite{shu2022test} by 3.81\%, DiffTPT~\cite{feng2023diverse} by 2.36\%), cache-based TDA~\cite{karmanov2024efficient} by 1.49\%, and prototype-based methods (DMN-ZS~\cite{zhang2024dual0} and DPE~\cite{zhang2024dual} by 1.10\% and 0.26\%, respectively). A similar trend was observed with the ViT-B/16 backbone, where BPRE outperforms TPT by 4.06\%, DiffTPT by 4.22\%, TDA by 1.49\%, DMN-ZS by 1.01\%, and DPE by 0.57\%. Notably, BPRE achieved the largest margins on the most challenging OOD datasets, such as ImageNet-A and ImageNet-Sketch, where the distribution gap is the greatest. These improvements were particularly meaningful given the strong performance of existing baselines and the inherent challenge of TTA without access to source data or target labels. The consistent gains across OOD benchmarks indicate that the bidirectional prototype-reward co-evolution mechanism is highly effective in identifying high-quality features and dynamically refining class prototypes, thereby achieving better semantic alignment with the target domain even under severe distribution shifts. Integrating similarity, confidence, and diversity within its multi-dimensional reward module, BPRE mitigated over-reliance on text priors and reduced misclassification, resulting in cross-modal misalignment. This positive feedback loop between rewards and prototypes enables progressive knowledge refinement, resulting in not only higher accuracy but also more stable and reliable performance across diverse scenarios. Overall, these results validated the robustness and generalizability of BPRE when domain shift is prevalent.

\begin{table*}[t]

\caption{Comparison with DPE on the SigLIP across 10 datasets.}
\renewcommand\arraystretch{1}
\renewcommand{\tabcolsep}{2pt}
\vspace{5pt}
\setlength{\abovecaptionskip}{2pt}  
\setlength{\belowcaptionskip}{2pt}  
\centering
\resizebox{0.80\linewidth}{!}{
  \begin{tabular}{lccccccccccc}
    \toprule
    Method & Aircraft & Caltech & Cars & DTD & EuroSAT & Flower & Food101 & Pets & SUN397 & UCF101 & Average \\
    \midrule
    DPE (SigLIP) & 10.44 & 62.39 & 14.76 & 61.88 & \textbf{38.58} & 72.68 & 26.18 & 79.34 & 66.96 & 55.30 & 48.85 \\
    \textbf{Ours (SigLIP)} & \textbf{13.82} & \textbf{64.26} & \textbf{15.83} & \textbf{62.58} & 38.40 & \textbf{73.04} & \textbf{41.83} & \textbf{79.37} & \textbf{67.35} & \textbf{52.74} & \textbf{50.92} \\
    \bottomrule
  \end{tabular}
}

\label{tab:siglip}
\vspace{-20pt}
\end{table*}

\subsubsection{Cross-Datasets Generalization} 

To assess the cross-dataset generalization capability of BPRE, we conducted experiments on 10 diverse benchmarks using two backbone architectures. As shown in Tab.~\ref{tab:fine-grained}, BPRE consistently achieved superior performance compared to existing methods. BPRE with ResNet-50 achieves significant improvements over the CLIP model (+7.52\%) and consistently outperforms state-of-the-art methods, including TPT (+5.68\%), DiffTPT (+3.49\%), TDA (+2.31\%), and DPE (+1.41\%). The performance gains are particularly notable on challenging fine-grained datasets, with substantial improvements on Aircraft (+2.25\%) and DTD (+2.78\%), where subtle visual distinctions between classes require more precise prototype representations.
For ViT-B/16, BPRE established a new SOTA benchmark with 70.10\% average accuracy, demonstrating consistent advantages over leading competitors DPE (+0.70\%) and TDA (+2.57\%). The method exhibits remarkable domain adaptability, showing significant improvements on specialized datasets such as Flower (+1.79\%) and Pets (+1.34\%). BPRE maintained strong performance on several datasets with varying numbers of classes, image resolutions, and visual features. Unlike the OOD experiments on ImageNet variants that primarily assess robustness to broad distribution shifts, fine-grained benchmarks highlight the need to distinguish subtle and localized visual differences between similar categories. The consistent superiority of BPRE in these tasks stems from its bidirectional prototype-reward co-evolution mechanism, which iteratively refines class prototypes using multi-dimensional feature quality assessment, integrating similarity, confidence, and diversity to select features that are both representative and discriminative. Our method is particularly effective where class boundaries are ambiguous and other methods may lack sufficient granularity. Furthermore, BPRE features a lightweight, training-free design that enables efficient adaptation without requiring additional data or complex optimization. Overall, these results demonstrated our effectiveness and generalizability for the unique challenges of fine-grained visual tasks.


\begin{table}[t]
\centering
\caption{Ablation studies of different components of the MQRM on the Food and ImageNet datasets.}
\label{tab:ablation-component}
\renewcommand\arraystretch{1.2}
\renewcommand{\tabcolsep}{5pt}
\small
\begin{tabular}{c|ccc|c|c}
\toprule
\multirow{2}{*}{\#} & \multicolumn{3}{c|}{Components} & Food & ImageNet \\
& $R_{\text{sim}}$ & $R_{\text{conf}}$ & $R_{\text{div}}$ & Acc. (\%) & Acc. (\%) \\
\midrule
1 & \ding{51} & \ding{55} & \ding{55} & 76.12 & 63.56 \\
2 & \ding{55} & \ding{51} & \ding{55} & 75.97 & 63.59 \\
3 & \ding{55} & \ding{55} & \ding{51} & 74.21 & 63.52 \\
4 & \ding{51} & \ding{51} & \ding{55} & 76.68 & 63.61 \\
5 & \ding{51} & \ding{55} & \ding{51} & 75.78 & 63.65 \\
6 & \ding{55} & \ding{51} & \ding{51} & 74.42 & 63.64 \\
\midrule
7 & \ding{51} & \ding{51} & \ding{51} & \textbf{77.59} & \textbf{63.77} \\
\bottomrule
\end{tabular}
\vspace{-5pt}
\end{table}

\subsubsection{Efficiency Comparison.} 

To evaluate the efficiency of BPRE, we compared its efficiency and accuracy with state-of-the-art test-time adaptation (TTA) methods on 50,000 ImageNet test samples. As shown in Tab.~\ref{table:efficiency}, while TPT and DiffTPT achieved modest gains (+0.93\% and +0.99\%), their high computational costs (>10h and >20h) make them impractical for practical use. In contrast, BPRE achieves an optimal balance between efficiency and performance, requiring only slightly more computation than DPE (3h 04min vs. 2h 37min) while delivering higher accuracy (+0.36\%). These results demonstrate that our bidirectional co-evolution mechanism enhances model generalization without significant computational overhead, making BPRE well-suited for test-time adaptation scenarios where both accuracy and efficiency are crucial.

\subsection{Ablation Studies}

\subsubsection{Robustness Analysis on Different Architectures}

To further assess the robustness and generalizability of our method, we evaluated it on the SigLIP architecture and compared it with DPE on 10 diverse datasets, as shown in Tab.~\ref{tab:siglip}. Our approach consistently surpassed DPE (SigLIP) on nearly all benchmarks, with notable improvements on datasets such as Aircraft (13.82\% vs. 10.44\%), Food101 (41.83\% vs. 26.18\%), and Caltech (64.26\% vs. 62.39\%). Our method achieved a 2.07\% higher accuracy (50.92\% vs. 48.85\%) on average. These results demonstrated that our approach maintained strong performance across a range of visual domains and dataset characteristics, including both fine-grained and large-scale recognition tasks. The consistent gains also indicated that our method was compatible with different backbone architectures, such as SigLIP, and was not limited by specific model designs. These findings further demonstrate the robustness and versatility of our approach in adapting to various architectural settings and data distributions.

\begin{table}[t]
\centering
\caption{Ablation study on different values of $M$.}
\label{tab:ablation-M}
\setlength\tabcolsep{4pt}
\small
\begin{tabular}{lccccc}
\toprule
$M$    & 1   &  2  &  \textbf{3}  & 4 & 5 \\
\midrule
ImageNet Acc. (\%)   &63.31 &63.52 & \textbf{63.77} &63.56 &63.72 \\
Aircraft Acc. (\%)   & 19.56 & 21.77 & \textbf{22.05} & 22.01 & 20.42\\
\bottomrule
\end{tabular}
\vspace{-5pt}
\end{table}

\subsubsection{Effectiveness of the MQRM}

We conducted an ablation study on Food101~\cite{bossard2014food} to evaluate each component in our Multi-dimensional Quality-aware Reward Module. As shown in Tab.~\ref{tab:ablation-component}, removing any reward component consistently degrades performance. Among single-component experiments, $R_{sim}$ (76.12\%) provides the strongest signal, followed by $R_{conf}$ (75.97\%) and $R_{div}$ (74.21\%). Dual-component configurations show improved results, with $R_{sim}$+$R_{conf}$ achieving the highest accuracy (76.68\%). The full implementation delivered superior performance (77.59\%), surpassing the best dual-component configuration by 0.91\%. This progressive improvement demonstrated the complementary nature of our reward components, where each addresses a distinct aspect of feature quality assessment. The observed results were attributed to the specific roles each reward component plays in the adaptation process. The similarity reward ($R_{sim}$) ensures that selected features are closely aligned with class prototypes, providing a strong foundation for semantic consistency. The confidence reward ($R_{conf}$) favors features with high prediction certainty, which helps reduce the influence of noisy or ambiguous samples. The diversity reward ($R_{div}$) encourages the selection of a broader range of informative features, preventing overfitting to a narrow subset of the data. When used in combination, these components not only reinforced strengths but also compensated for their individual limitations. For instance, while $R_{sim}$ alone may overlook uncertain yet informative features, adding $R_{conf}$ and $R_{div}$ ensures that the model remains robust to uncertainty and captures a richer set of discriminative cues. The best performance achieved by the full reward module highlights the importance of a holistic approach to feature quality assessment, enabling more effective and stable prototype refinement during test-time adaptation.

\begin{table}[t]
\centering
\caption{Ablation study on different values of $R_{\text{min}}$.}
\label{tab:ablation-Rmin}
\setlength\tabcolsep{4pt}
\small
\begin{tabular}{lccccc}
\toprule
$R_{\text{min}}$ & 0.0 & 0.05 & \textbf{0.1} & 0.2 & 0.4 \\
\midrule
ImageNet Acc. (\%) & 62.76 & 63.54 & \textbf{63.77} & 63.68 & 63.59 \\
Food Acc. (\%) & 76.31 & 76.89 & \textbf{77.59} & 77.37 & 77.03 \\

\bottomrule
\end{tabular}
\vspace{-5pt}
\end{table}

\subsubsection{Ablation studies on different values of queue size $M$}

\begin{figure}[t]
\centering
\includegraphics[width=0.80\columnwidth]{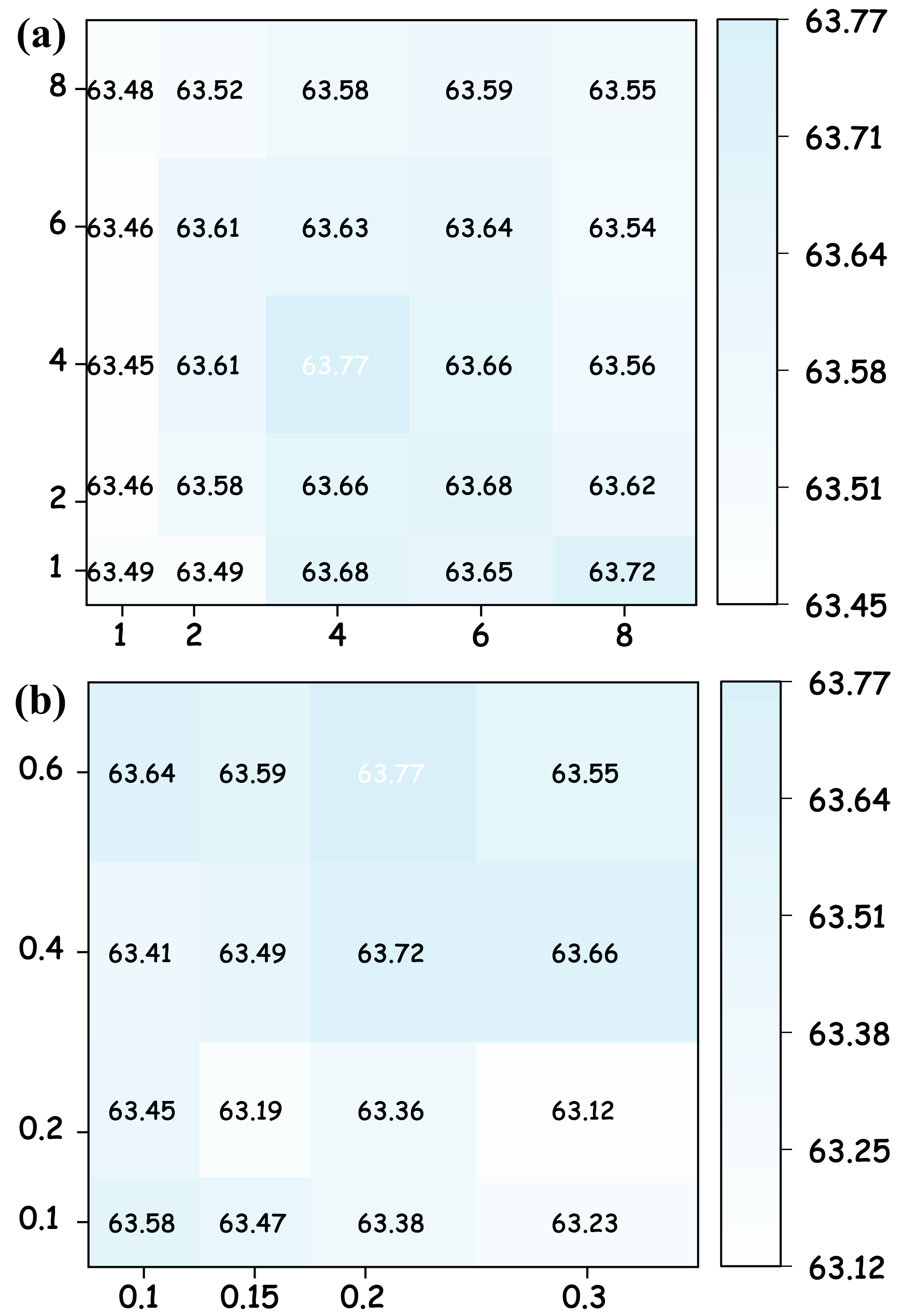}
\caption{Hyperparameter analysis of $\alpha$, $\beta$, $\lambda_{\text{sim}}$, and $\lambda_{\text{conf}}$ was conducted on the ImageNet dataset using a ResNet-50 backbone.}
\label{fig:4}
\vspace{-5pt}
\end{figure}

To investigate the impact of the feature memory queue size $M$ on recognition performance, we conducted ablation studies on both the ImageNet and Aircraft datasets. Tab.~\ref{tab:ablation-M} presents the results of our ablation study on the feature memory queue size $M$ for both ImageNet and Aircraft datasets. The findings reveal a consistent trend across datasets: accuracy increases as $M$ grows, reaching its peak at $M=3$ (63.77\% on ImageNet and 22.05\% on Aircraft), after which performance begins to decline. This pattern underscores the importance of appropriately setting the memory queue size to balance feature diversity and noise accumulation. When $M$ is too small (e.g., $M=1$ or $2$), the queue contains only the most recent features, limiting the diversity of prototype updates and making the model susceptible to short-term fluctuations or outliers in the data stream. This restricted memory capacity hinders the model’s ability to generalize and accurately capture the underlying class distributions, resulting in suboptimal prototype refinement and lower accuracy on both datasets. As $M$ increases, the queue retains a more diverse set of historical features, which enhances the representativeness of prototypes and leads to improved recognition performance. However, when $M$ becomes too large (e.g., $M=4$ or $5$), the queue starts to accumulate older or potentially irrelevant features that may no longer reflect the current data distribution. This introduces noise, dilutes the most salient and up-to-date class features, and ultimately degrades both prototype quality and recognition accuracy. The optimal value at $M=3$ thus reflects a critical balance: it provides sufficient diversity for robust prototype evolution while avoiding the adverse effects of excessive noise from outdated features. This observation highlights the necessity of carefully calibrating the historical feature retention mechanism to ensure stable and effective test-time adaptation across different domains. Moreover, it demonstrates that robust domain adaptation depends not only on the quality of feature selection but also on the temporal scope of memory used for prototype updates.

\subsubsection{Ablation studies on different values of $R_{min}$}

To evaluate the effect of the minimum reward threshold $R_{min}$ on model performance, we conducted ablation studies on both the ImageNet and Food datasets. Tab.~\ref{tab:ablation-Rmin} presents our ablation study on the minimum reward threshold $R_{min}$ for both ImageNet and Food datasets. The results consistently show that setting $R_{min}=0.1$ yields the highest accuracy (63.77\% on ImageNet and 77.59\% on Food). When $R_{min}$ is set too low (e.g., $0.0$ or $0.05$), a large number of low-quality features are incorporated into prototype updates. This inclusion introduces noise, weakens the discriminative power of class prototypes, and ultimately leads to suboptimal recognition performance on both datasets. Conversely, a higher threshold (e.g., $0.2$ or $0.4$) enforces stricter filtering, ensuring that only high-quality features contribute to prototype updates. While this increases feature quality, it also reduces feature diversity and risks overfitting to a limited subset of the data, thereby impairing the model’s adaptability and generalization to new samples. The optimal performance observed at $R_{min}=0.1$ suggests that a moderate threshold effectively balances feature quality and diversity. This setting allows the model to leverage informative features while filtering out detrimental noise, resulting in more representative and robust prototypes. These findings highlight the importance of carefully tuning $R_{min}$ to maintain an optimal trade-off between strictness and inclusiveness in feature selection, which is crucial for stable and effective test-time adaptation across diverse domains.

\subsubsection{More Sensitivity Analyses of Hyper-Parameters}

To further investigate the influence of key hyperparameters on model performance, we conducted comprehensive sensitivity analyses on the ImageNet dataset, as illustrated in Fig.~\ref{fig:4}. As shown in Fig.~\ref{fig:4}(a), model accuracy is maximized when both temperature parameters $\alpha$ and $\beta$ are set to 4, indicating that an appropriate sharpness in the reward distribution and a balanced prototype evolution rate are critical for stable adaptation. Deviating from these values either overly smooths or over-concentrates updates, leading to suboptimal learning dynamics. Fig.~\ref{fig:4}(b) further investigates the impact of reward component weighting. The best performance was achieved with $\lambda_{sim}=0.6$, $\lambda_{conf}=0.2$, and $\lambda_{div}=0.2$. This setting prioritizes semantic similarity in the reward calculation, ensuring prototypes remain closely aligned with class semantics, while still leveraging confidence and diversity to avoid overfitting and encourage robust representation. These results validated that our design emphasized similarity while maintaining contributions from complementary metrics, highlighting the significance of carefully balancing different reward components for effective prototype evolution and TTA.

\section{Conclusion}

We proposed BPRE, a novel test-time adaptation framework for VLMs that addresses distribution shifts through bidirectional prototype-reward co-evolution. Our MQRM joins semantic similarity, confidence, and diversity metrics to comprehensively assess feature quality, thereby reducing the dependency of our method on potentially biased text representations during adaptation. The Prototype-Reward co-Evolution mechanism establishes a self-reinforcing cycle where refined prototypes enhance reward estimation and improved rewards guide prototype updates. Extensive experiments on 15 diverse datasets demonstrate that BPRE achieved superior performance compared to SOTA methods in both natural distribution shifts and cross-dataset generalization scenarios. These results highlight the effectiveness of our bidirectional co-evolution approach in enabling robust adaptation to domain shifts without requiring source data or target labels, positioning it as a promising solution for vision-language model (VLM) deployment.


\bibliographystyle{IEEEtran}
\bibliography{references}

\end{document}